\documentclass[runningheads]{llncs}

\usepackage[mobile]{eccv}

\usepackage{eccvabbrv}

\usepackage{epsfig}

\usepackage{graphicx}
\usepackage{booktabs}
\usepackage{multirow}
\usepackage{epigraph}
\usepackage{listings}
\usepackage{xcolor}
\usepackage[normalem]{ulem}
\useunder{\uline}{\ul}{}

\usepackage{float}

\usepackage[accsupp]{axessibility}  

\usepackage{hyperref}

\usepackage{orcidlink}

\definecolor{codegreen}{rgb}{0,0.6,0}
\definecolor{codegray}{rgb}{0.5,0.5,0.5}
\definecolor{codepurple}{rgb}{0.58,0,0.82}
\definecolor{backcolour}{rgb}{0.95,0.95,0.92}

\lstdefinestyle{mystyle}{
    backgroundcolor=\color{backcolour},   
    commentstyle=\color{codegreen},
    keywordstyle=\color{magenta},
    numberstyle=\tiny\color{codegray},
    stringstyle=\color{codepurple},
    basicstyle=\ttfamily\footnotesize,
    breakatwhitespace=false,         
    breaklines=true,                 
    captionpos=b,                    
    keepspaces=true,                 
    numbers=left,                    
    numbersep=5pt,                  
    showspaces=false,                
    showstringspaces=false,
    showtabs=false,                  
    tabsize=2
}

\begin{document}

\title{Seeing Faces in Things:\\A Model and Dataset for Pareidolia} 


\author{
Mark Hamilton\inst{1,2}\orcidlink{0000-0002-0260-0101} \and
Simon Stent\inst{3}\orcidlink{0000-0002-2623-6383}\and
Vasha DuTell\inst{1}\orcidlink{0000-0001-8625-1350} \and
Anne Harrington \inst{1}\orcidlink{0009-0000-9441-2687}\and
Jennifer Corbett \inst{1}\orcidlink{0000-0002-9412-7963}\and
Ruth Rosenholtz\inst{4}\orcidlink{0000-0001-5299-0331} \and
William T. Freeman\inst{1}\orcidlink{0000-0002-2231-7995}
}

\authorrunning{M.~Hamilton, S.~Stent, V.~DuTell, et al.}

\institute{MIT \and Microsoft\and 
Toyota Research Institute \and
NVIDIA}

\maketitle

\begin{abstract}
The human visual system is well-tuned to detect faces of all shapes and sizes. While this brings obvious survival advantages, such as a better chance of spotting unknown predators in the bush, it also leads to spurious face detections. ``Face pareidolia'' describes the perception of face-like structure among otherwise random stimuli: seeing faces in coffee stains or clouds in the sky. In this paper, we study face pareidolia from a computer vision perspective. We present an image dataset of ``Faces in Things'', consisting of five thousand web images with human-annotated pareidolic faces. Using this dataset, we examine the extent to which a state-of-the-art human face detector exhibits pareidolia, and find a significant behavioral gap between humans and machines. We find that the evolutionary need for humans to detect animal faces, as well as human faces, may explain some of this gap. Finally, we propose a simple statistical model of pareidolia in images. Through studies on human subjects and our pareidolic face detectors we confirm a key prediction of our model regarding what image conditions are most likely to induce pareidolia. Dataset and Website: \href{https://aka.ms/faces-in-things}{https://aka.ms/faces-in-things}
\end{abstract}

\section{Introduction}
\label{sec:intro}

\epigraph{\textbf{Hamlet}: Do you see yonder cloud that's almost in the shape of a camel?\\
\textbf{Polonius}: By the Mass and 'tis, like a camel indeed.\\
\textbf{Hamlet}: Methinks it is a weasel.\\
\textbf{Polonius}: It is back'd like a weasel.\\
\textbf{Hamlet}: Or like a whale.\\
\textbf{Polonius}: Very like a whale.}{--- \textit{Hamlet, Act III, Scene ii}, William Shakespeare}

\begin{figure}[ht]
    \begin{minipage}{0.4\textwidth}
        \centering
        \includegraphics[width=\linewidth]{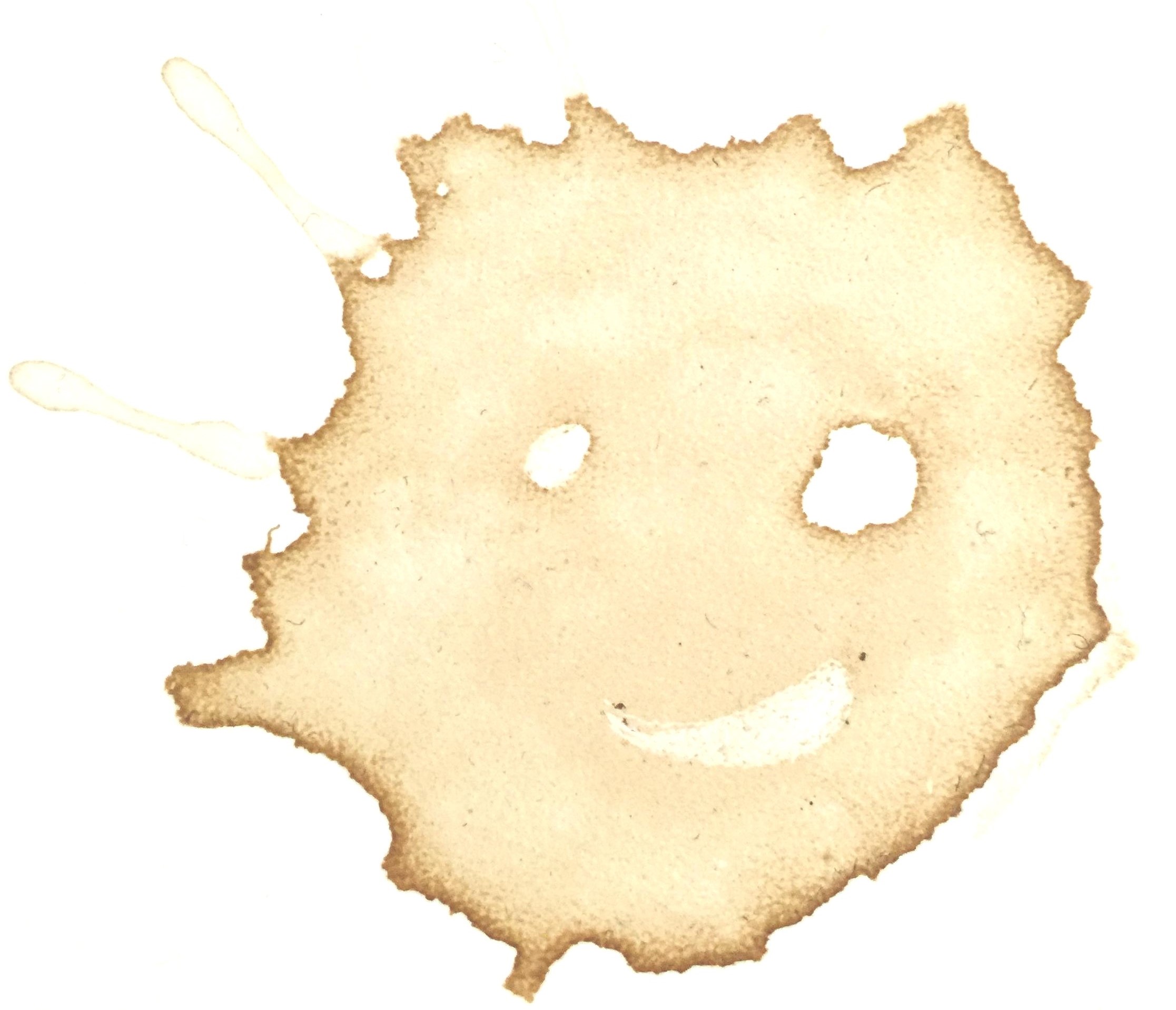}
        \label{fig:pull}
    \end{minipage}%
    \begin{minipage}{0.6\textwidth}
        \caption{You print out an exciting new computer vision paper to review, but as you sit down at your desk to start reading you knock over your coffee cup. At first, you are annoyed, but then, you laugh! The sight of the stain induces ``pareidolia'' in your brain: rather than an unsightly blemish, you see a happy face. In this paper we explore the phenomenon of face pareidolia: Why don't we see faces all the time? Why do we see them at all when they are clearly so different from human faces? Can a better understanding of face pareidolia help computer vision--based face detection?}
    \end{minipage}
\end{figure}

Pareidolia is a type of visual ``apophenia'', which refers to the perception of patterns in random data.
This occurs frequently in human perception as we look at clouds, mountain skylines, and burnt toast.
Pareidolia is even described in an exchange in Hamlet~\cite{Shakespeare1954}.
When it was first described, pareidolia was seen as an early symptom of psychosis~\cite{Conrad1958,sims1988symptoms}.
Today we know pareidolia is common among healthy humans~\cite{summerfield2006mistaking} and infants~\cite{kato_pareidolia_2015}.
It is also not confined to humans: rhesus macaques, for example, have been shown to spend more time fixating on pareidolic than non-pareidolic images, in a manner similar to humans~\cite{taubert_face_2017}.

As an intriguing phenomenon of our visual system, pareidolia presents many opportunities in the study of the visual perception of both humans and machines.
It offers a controlled setting in which to study object detection: we can present random signals to the visual system and study what detections arise.
Do computer vision detectors exhibit similar misidentifications, and if not, why not?
Why don't humans see pareidolic effects everywhere, in any textured region?


To help answer these questions, we introduce an annotated dataset of five thousand pareidolic face images, called ``Faces in Things''.
With this dataset, we examine whether modern computer vision face detection systems, trained to robustly detect human faces, exhibit pareidolia.
We show that a state-of-the-art neural network trained on the popular WIDER FACE detection benchmark~\cite{yang_wider_2016} fails to detect pareidolic faces well, even when detection thresholds are relaxed.
By fine-tuning the same model on the Faces in Things training data we create a simple and strong baseline for the task of pareidolic face detection, which shows that significantly higher machine pareidolic performance is within reach.

Next, we explore how we might bridge this gap to supervised---or ultimately, human---performance, without access to pareidolic training data? Could pareidolia appear in a face detector in a more natural way? The Faces in Things dataset provides a clean testbed to explore these questions in machines. We test a variety of different interventions ranging from image augmentation techniques to additional sources of training data. We find one possible mechanism that accounts for roughly half of the performance gap: when models are fine-tuned to detect \textit{animal} faces, pareidolic face detection is significantly improved. This suggests that face pareidolia may arise in part from a more general, evolutionary need to detect diverse faces in the natural environment.

Finally, we consider why pareidolic faces are not all around us, and why certain textures seem to cause the effect more often. 
We propose two simple mathematical models, a simple Gaussian process model, and a second deep feature-based model, that capture important features of pareidolia.
In particular, we show how these simple models both predict a ``Goldilocks'' zone, where conditions are ideal to induce pareidolia.
We confirm the existence of this zone with experiments on both human subjects and face detection models.

Through the contributions of our open-source dataset, models, and experimental findings, we bring the study of the intriguing phenomenon of face pareidolia to the computer vision community. 

\section{Related Work}
\label{sec:related}

\textbf{Face detection.}
One of the most famous early examples of face detection was the Viola-Jones face detector~\cite{viola_rapid_2001,viola_robust_2004}.
This detector used binary Haar features through simple-to-compute integral images and  achieved greater precision and efficiency than early neural-network-based detectors~\cite{rowley_human_1996,rowley_neural_1998} and other feature-based methods~\cite{ming-hsuan_yang_detecting_2002,rein-lien_hsu_face_2002}.
Following the deep learning breakthroughs of the 2010s, methods transitioned from hand-crafted features to learned features, and convolutional neural networks (CNNs) achieved close to human levels of performance on ever-larger datasets~\cite{deng_retinaface_2019,ferrari_pyramidbox_2018,liao_fast_2016,li_convolutional_2015,li_dsfd_2019,mathias_face_2014,ranjan_deep_2018,zhang_s3fd_2017}.
For a broader survey of face detection methods,
we refer the interested reader to~\cite{zafeiriou_survey_2015,ranjan_deep_2018}.
In our work, we use the recent RetinaFace model~\cite{deng_retinaface_2019} as a strong face detection baseline.

\textbf{Neuroscience of face pareidolia.}
The face is a highly unique stimulus for the human visual system~\cite{tsao2008mechanisms,leopold2010comparative}: we find faces easy to spot and difficult to ignore. Face detection can occur in both noise and highly degraded images \cite{campbell1983much}. Prior work shows that face detection occurs in a dedicated brain region, the Fusiform Face Area ~\cite{mcgugin_high-resolution_2012}.
But exactly what constitutes a face for the visual cortex and what are the mechanisms underlying pareidolia? A recent study into the temporal dynamics of neuro-imaging data during pareidolic face viewing showed results consistent with ``a broadly-tuned face detection mechanism that privileges sensitivity over selectivity''~\cite{wardle2020rapid}.
Pareidolic faces do more than give the impression of the presence of faces: 
\cite{takahashi_gaze_2013} show that they can trigger an additional face-specific attentional process, consuming more time and processing power than similar non-pareidolic stimuli, and even enhancing the detection of face-pareidolic objects~\cite{takahashi_seeing_2015}.
Analyses in~\cite{liu_seeing_2014} revealed a network of neurons in the brain specialized to detect face pareidolia. 
Their results suggested that face processing has a strong top-down component whereby sensory input with even the slightest suggestion of a face can result in the interpretation of a face.
Such top-down information might be supportive of some form of inverse rendering as a cognitive mechanism to explain the remarkable robustness of human perception of faces in degraded viewing conditions~\cite{egger2020inverse}. 
While our dataset allows the study of several types of face detection models, we focus our study on feed-forward neural networks which are known to yield close to human performance on challenging ``in-the-wild'' datasets~\cite{yang_wider_2016}.

\begin{figure*}[t]
    \centering
    \includegraphics[width=\columnwidth]{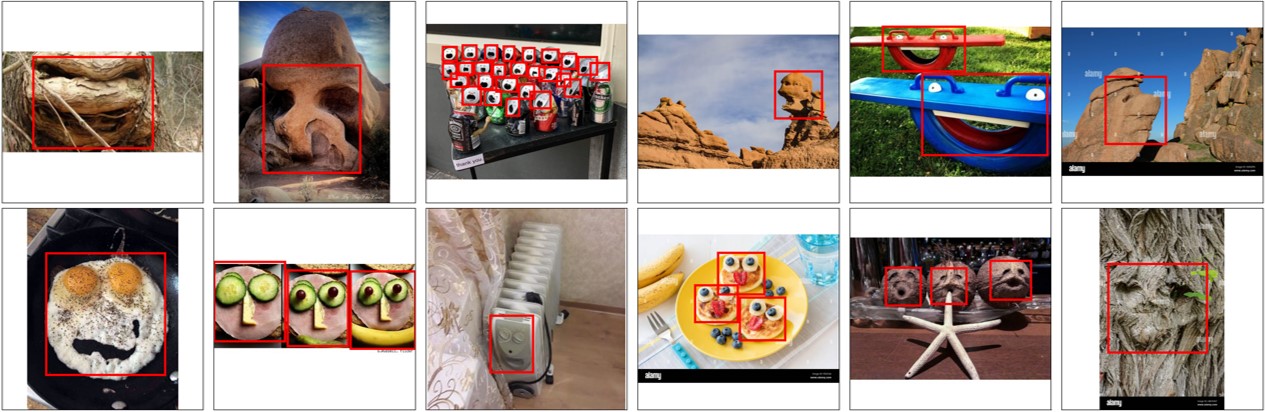}
    \caption{\textbf{Examples of face pareidolia from our ``Faces in Things'' dataset.} Faces in Things consists of five thousand images annotated with bounding boxes (shown here), and facial attributes such as perceived emotion, gender, and intentionality.}
    \label{fig:example_images}
\end{figure*}

\textbf{Face pareidolia in computer vision.}
Face detection and face recognition have been core topics in computer vision for many decades, but the study of face pareidolia---and its deep relationship with visual object representation learning---has been relatively overlooked. 
Face pareidolia has some similarities with the problem of cross-modal recognition or cross-depiction: recognizing the same objects across different modalities irrespective of how the object is visually depicted.
This has been explored particularly in the context of detecting faces, people and objects across modalities such as photography, different art movements, cartoons and sketches~\cite{ginosar_detecting_2014,cai_cross-depiction_2015,MishraECCV16,westlake_detecting_2016}. The importance of capturing spatial relationships for robust cross-modal detection has also been highlighted~\cite{cai_cross-depiction_2015}.
The work of Castrejon \textit{et al.}~\cite{castrejon_learning_2016} showed how, when learning cross-modal scene representations with neural networks, units would emerge in the shared representation that tended to activate on consistent concepts, independently of the modality.
This tendency was used by Abbas \& Chalup~\cite{abbas_face_2019}, who found that mid-level units learned during human face detection could generalize to detect semantically similar facial key-points in pareidolic images, showing promise for pareidolia to emerge. However, the authors only evaluated the method qualitatively over a small test set of ten images.
One route to a larger dataset may be through pareidolic image generation, which shows promise but does not yet produce convincingly natural images~\cite{endo2024systematic}.
Curating a larger dataset, ``Totally-Looks-Like''~\cite{rosenfeld2018totally} explored the perceptual judgment of image similarity between humans and CNNs, using images which had been paired by humans as visually similar but semantically disparate. They found that visual representations extracted from CNNs such as ResNet~\cite{he2016deep} perform poorly in terms of reproducing the matching selected by humans. Though this dataset is of similar size to ours (6k samples), it is not specifically tailed towards face pareidolia and offers no bounding box or key-point annotations. 
Other datasets such as COCO-Periph~\cite{harringtoncoco} have been used to show that object detection behavior in CNNs and transformers diverge from human perception in peripheral vision. 

In summary, there has yet to be a computational model of how or why pareidolia might arise proposed in the literature.
Moreover, despite the abundance of face detection datasets~\cite{ranjan_deep_2018}, there is no large-scale dataset to directly support the study of face pareidolia. 
A large-scale pareidolia dataset would help the community explore the mechanisms underlying pareidolia, which may in turn help us to understand and harness human visual attention (which is drawn towards face-like objects), reduce pareidolic false positives in face detectors, help designers avoid or create pareidolia, improve pareidolic animation, and create systems that better understand how humans perceive the world.

\section{Faces in Things Dataset}
\begin{figure}[t]
    \centering
    \includegraphics[width=0.8\columnwidth]{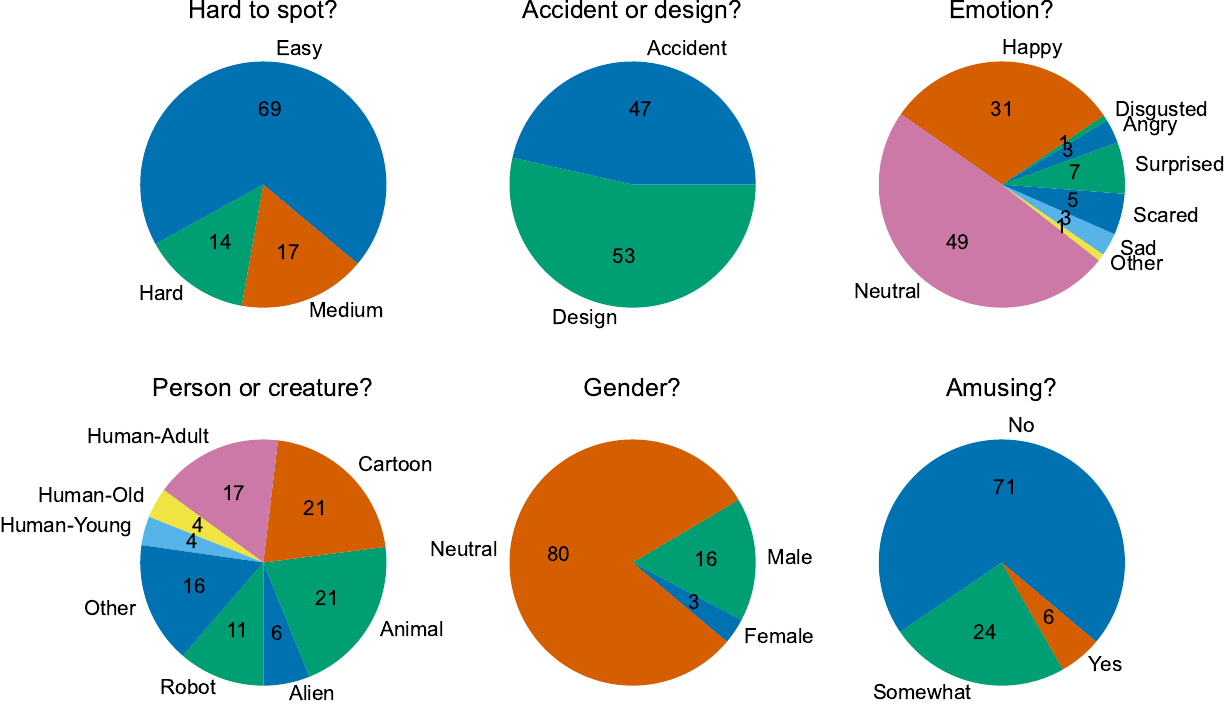}
    \caption{\textbf{Attributes of the Faces in Things Dataset.} We find that 31\% of faces are considered challenging to spot; faces are largely (31\%) judged as happy; approximately half (47\%) are judged as accidental rather than by design; animals and humans are seen in roughly equal numbers; and we observe a slight bias (16\% vs 3\%) towards male over female faces, similar to biases observed in prior studies~\cite{wardle22,wardle23cognition}.}
    \label{fig:dataset_stats}
\end{figure}

To address this gap, we begin by sampling candidate pareidolic images from the LAION-5B dataset~\cite{schuhmann2022laionb}.
This dataset consists of 5.85 billion CLIP-filtered image-text pairs, of which 40\% of captions contain English.
We use CLIP retrieval~\cite{beaumont-2022-clip-retrieval} to build a raw image set based on text queries including ``pareidolia'', ``faces in things'', ``accidental faces'', and ``[object] looks like a face''.
We download images, check for duplicates, then downsample to $512\times512$ pixels while preserving the aspect ratio with white-space padding. 
We used the VGG Image Annotation tool~\cite{dutta2019vgg} to manually annotate images, removing samples that contain the faces of actual humans or animals.
Some examples of annotated images are shown in Fig.~\ref{fig:example_images}.
Our annotations include the bounding boxes of pareidolic faces and basic facial attributes as summarized in Fig.~\ref{fig:dataset_stats}.
Though beyond the scope of the current paper, we note that these attributes could be useful for other future studies.
We divide the dataset at random into training (70\%) and testing (30\%) sets.  We refer to this as the 'Pareidolic' dataset.

\section{Experiments}
\label{sec:exp}

\begin{figure}[t]
    \centering\includegraphics[width=0.6\columnwidth]{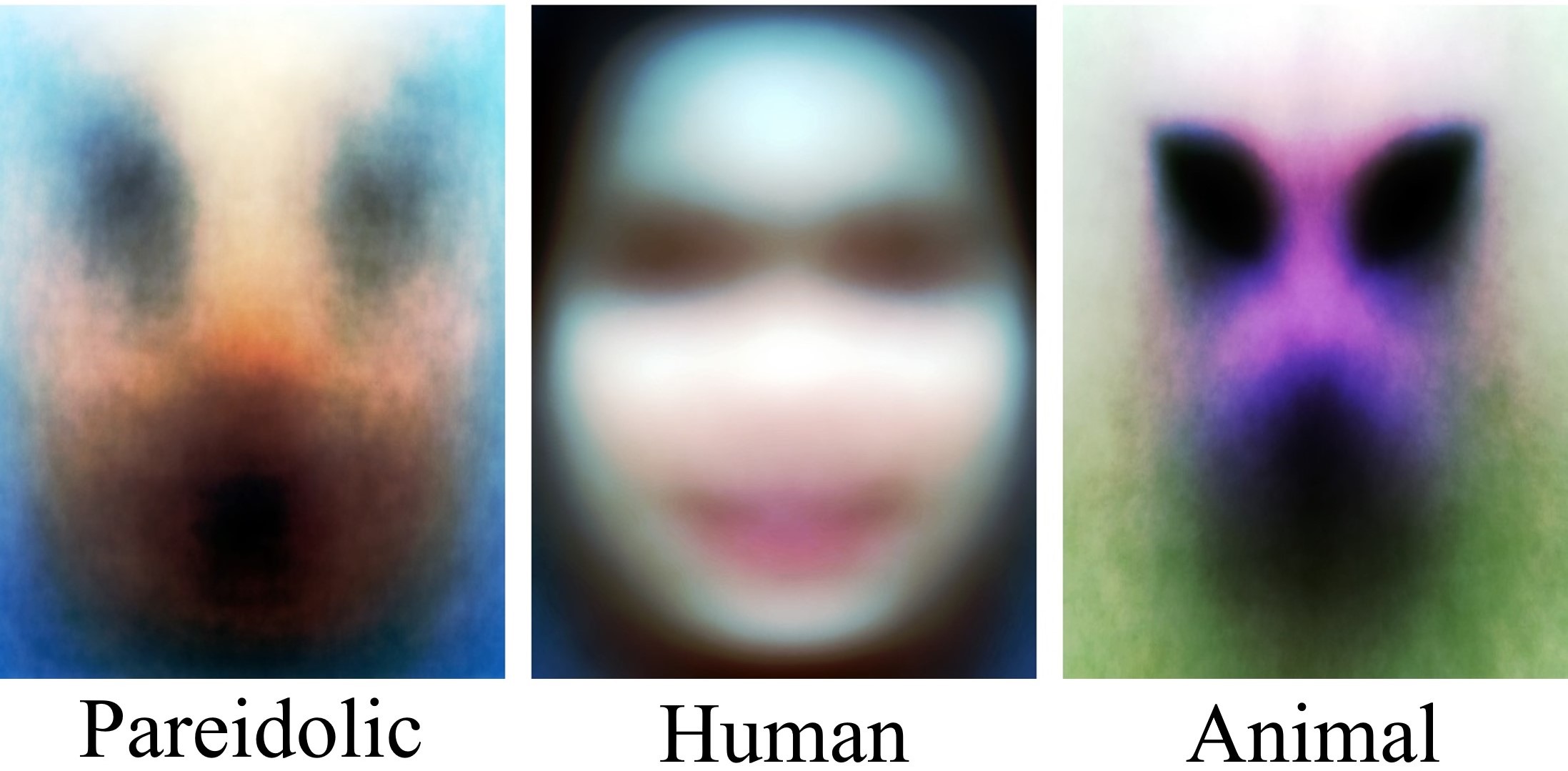}
    \caption{\textbf{The Appearance of an Average Pareidolic Face.} Per-channel histogram-equalized average images for registered pareidolic faces (our Faces in Things dataset), human faces (samples from the WIDER FACE dataset~\cite{yang_wider_2016}), and animal faces (AnimalWeb~\cite{khan2020animalweb}). The average pareidolic face, while less distinct than human or animal, has surprisingly clear eye, nose, and mouth features, and vertical symmetry.}
    \label{fig:average_faces}
\end{figure}

\paragraph{Datasets.}
We use the following additional datasets. Fig.~\ref{fig:average_faces} shows the average faces within our dataset (Pareidolic) and the WIDER FACE (Human), and AnimalWeb (Animal) datasets. 

\textbf{WIDER FACE}~\cite{yang_wider_2016} is a popular face detection benchmark dataset with 32,203 images and 393,703 faces. It contains a high degree of variability in scale, pose, makeup, lighting, emotion, and occlusion, organized across 61 event classes. We use the provided $40\%/10\%/50\%$ splits for training, validation, and testing. We refer to this as the `Human' dataset.

\textbf{AnimalWeb}~\cite{khan2020animalweb} is a collection of 22,451 faces from 334 diverse species and 21 animal orders across biological taxonomy. These faces are captured `in-the-wild' and are consistently annotated with 9 landmarks on key facial features. We convert these landmarks to bounding boxes, by finding the tightest box that captures the points and expanding this box's width and height by $15\%$.  We refer to this as the `Animal' dataset.

\textbf{WIDER FACE Corruptions.}
To measure whether pareidolia could arise from common data augmentations we corrupt the WIDER FACE images using the level 3 strength of the corruptions used in both the COCO-C~\cite{michaelis2019benchmarking} and ImageNet-C~\cite{hendrycks2019robustness} datasets. We also include a Sobel filtering corruption \cite{farid1997optimally} which has been shown to reduce a model's dependence on texture information \cite{ji2019invariant}.

\begin{figure*}[t]
    \centering
    \includegraphics[width=\columnwidth]{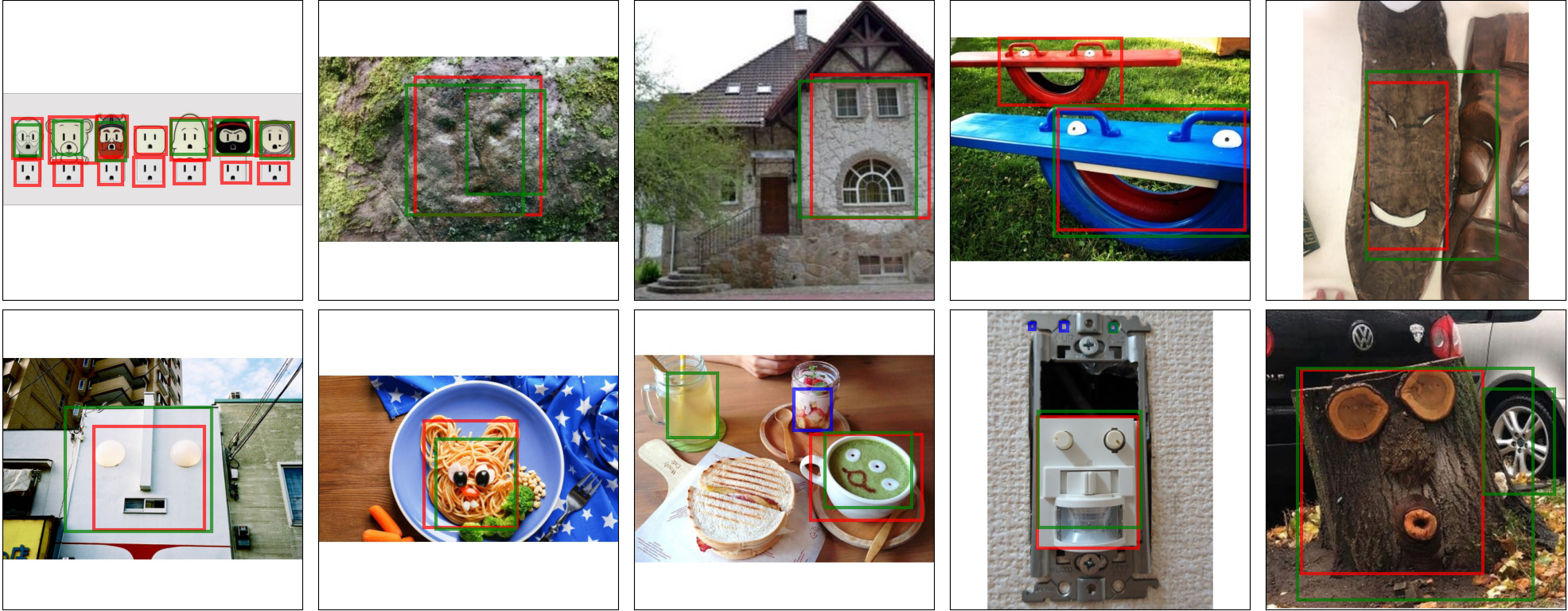}
    \caption{\textbf{Qualitative Analysis of Transfer Experiments.} On a sample of held-out test images, we visualize the confident $(p>10\%)$ detections of our \textcolor{red}{ground truth (red)}, our model \textcolor{blue}{fine-tuned on human faces (blue)}, and our model \textcolor{green}{fine-tuned on animal faces (green)}. It is evident from these and Table~\ref{tab:results_a2p} that fine-tuning on animal faces significantly boosts the model's ability to detect pareidolic faces.}
    \label{fig:qualitative}
\end{figure*}

\paragraph{Models and Training.}
We use RetinaFace~\cite{deng_retinaface_2019} which achieves state-of-the-art performance on WIDER FACE easy and medium subsets and is the third-best face detector on the hard subset, missing the top model by less than a percentage point of Average Precision (AP). We perform experiments using both their MobileNet~\cite{howard2017mobilenets} and ResNet50~\cite{he2016deep} backbones and use the Pytorch\_Retinaface~\cite{pytorch_retinaface} repository to ensure the same experimental conditions, dataset characteristics, and preprocessing. We use pre-trained models provided by this repository and fine-tune them for 10 epochs with the AdamW optimizer~\cite{loshchilov2017decoupled} using a learning rate of $10^{-4}$ and a weight decay of $5\times10^{-4}$. We verify that fine-tuning using this strategy on the original WIDER FACE training dataset does not hurt model performance. When fine-tuning on Faces in Things (Pareidolia), AnimalWeb (Animal), WIDER FACE (Human), and Corruption datasets we randomly replace images in the original WIDER FACE stream of training data with data from the target dataset $90\%$ of the time. This allows the network to learn the new task without catastrophic forgetting~\cite{kirkpatrick2017overcoming}. These changes to the optimizer, learning rate, number of epochs, and stream of training data are the only changes we make to the training paradigm of~\cite{deng_retinaface_2019}. Figures in this work use the MobileNet architecture of RetinaNet unless specified otherwise. AP evaluation computations share the same setting and parameters as~\cite{deng_retinaface_2019}.

\subsection{Does a SOTA Face Detector Exhibit Pareidolia?}
\label{sota}

We measure the Average Precision (AP) of the MobileNet and ResNet50 RetinaNet architectures on the Faces in Things dataset. The first row of Table~\ref{tab:results_a2p} shows results for existing pre-trained models, and the second row shows those for models fine-tuned on the original WIDER FACE training data. These act as control groups to ensure our transfer learning procedure does not interfere with our measurement of the effects of other interventions. Though these models exhibit pareidolia to a small extent, they fall far short of a model fine-tuned to detect pareidolic faces. Fig.~\ref{fig:qualitative} also depicts some of these predictions with blue boxes. On the whole, the models trained only on human faces are largely silent across the Faces in Things dataset.

\begin{figure*}[t]
    \centering
    \vspace{-.1in}
    \includegraphics[width=\columnwidth]{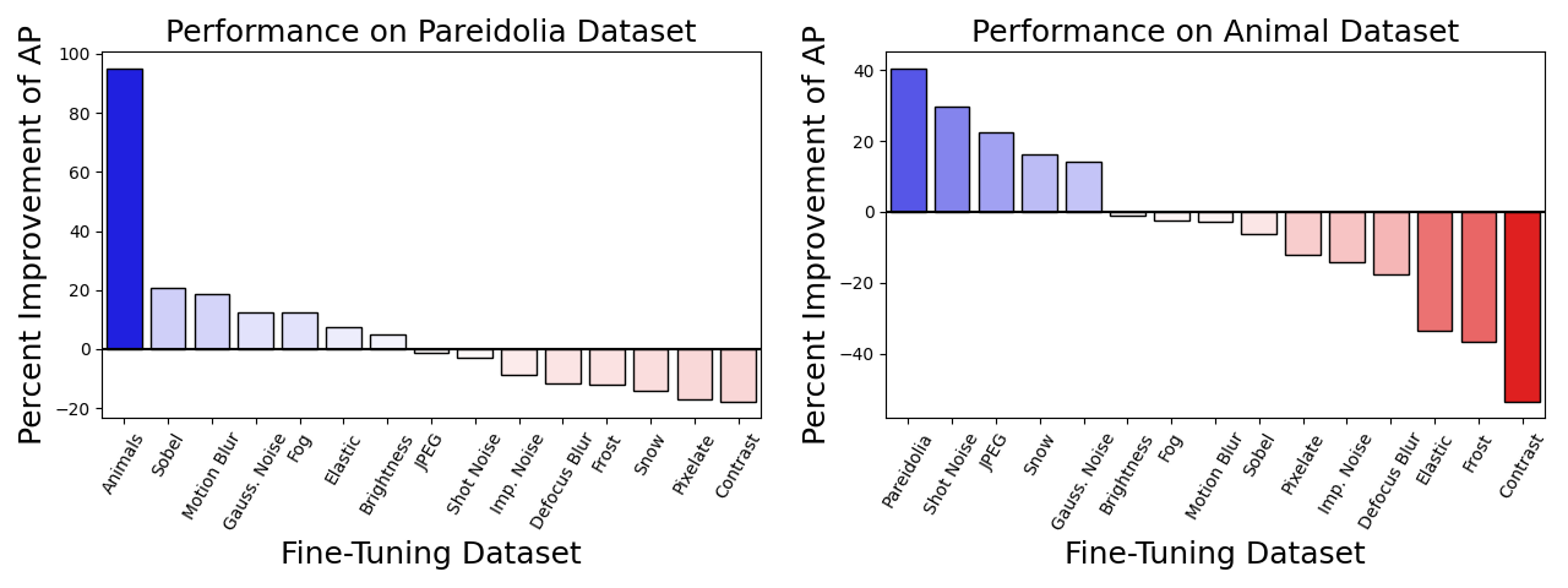}  
    \vspace{-.1in}
    \caption{\textbf{Measuring the effect of several training interventions on pareidolic face detection} The left plot shows that fine-tuning RetinaNet on animals improves pareidolic face detection more than any other intervention. Conversely, the right plot shows that pareidolic fine-tuning improves animal face detection performance.}
    \label{fig:transfer_learning}
\end{figure*}

\subsection{How Might Pareidolia Emerge?}

The WIDER FACE dataset is known for its diversity of lighting, pose, makeup, emotion, and scale of faces. This fact, coupled with the results of Section~\ref{sota} begs the question: What else do models need to experience pareidolia as humans do? The Faces in Things dataset provides a clean and robust setting to explore the development of pareidolia in algorithms. Unlike in humans, where it is impossible to causally intervene on their facial training data, we can easily modify an algorithm's training data. This makes it possible to explore whether one can induce pareidolia in algorithms using specific stimuli. 
 
To this end, we investigate whether a variety of training data interventions can induce pareidolia in algorithms. In particular, we measure the effect of adding several data augmentations from the COCO-C~\cite{michaelis2019benchmarking} and ImageNet-C~\cite{hendrycks2019robustness} datasets and explore a Sobel filtering augmentation which reduces models' dependence on texture. Additionally, we also measure the effect of adding animal faces to the training data. Animal faces show a far greater breadth of variation in coloration, structure, and appearance than human faces. Recognizing animal faces provides many evolutionary advantages including gaze detection during hunting and avoiding onlooking predators. The generality required to detect this wide space of faces could yield a greater number of ``false positives'' that lead to the sensation of pareidolia. Indeed, some recent studies provide some corroborating evidence for this hypothesis. Firstly, Rhesus Monkeys exhibit pareidolia~\cite{taubert_face_2017} showing this effect does not only occur in humans. Secondly, the experience of pareidolia is a rapid cognitive process and not a ``late re-interpretation'' of input signals~\cite{hadjikhani2009early,wardle2020rapid} which the authors conclude is evidence that pareidolia could be linked to the need to quickly react to predators. 

We plot the change to pareidolic face detection performance as a function of each training intervention in the left panel of Figure \ref{fig:transfer_learning}. Of the different corruption interventions, we find that Sobel filtering, motion blur, Gaussian noise, and fog tend to slightly improve pareidolic face detection while most other corruptions do not improve pareidolic face detection performance. Most strikingly, the addition of animal faces to the training data roughly doubles the algorithms' ability to detect pareidolic faces compared to the control group, closing around half of the gap between a human-trained model and a pareidolia trained model. Reciprocally, the right-hand plot of Figure \ref{fig:transfer_learning} shows that fine-tuning on pareidolic images yields the greatest improvement in animal face detection. We further explore this phenomenon in Table~\ref{tab:results_a2p}, where we show that this effect occurs across both MobileNet and ResNet50 architectures. Finally, we also show that adding a small number of animal faces (30\% animal 70\% pareidolic) can improve pareidolic face detection performance over pareidolic images alone.

\begin{table}[t]
\begin{centering}
\begin{tabular}{crr}
\toprule
\multirow{2}{*}{Finetuning} & \multicolumn{2}{c}{AP}            \\ \cline{2-3} 
                            & MobileNet       & ResNet50        \\ \midrule
None                        & 7.9\%           & 2.8\%           \\
Human (Control)             & 9.8\%           & 3.6\%           \\
Animal                      & 16.7\%          & 15.4\%          \\
Pareidolia                  & {\ul 33.9\%}    & {\ul 27.1\%}    \\
Animal + Pareidolia         & \textbf{36.4\%} & \textbf{31.7\%} \\ \bottomrule
\end{tabular}
\vspace{6pt}
\caption{\textbf{Effect of Fine Tuning on Pareidolic Face Detection.} Our results show that WIDER FACE-trained RetinaFace models do not detect many pareidolic images. Fine-tuning these models on animal faces approximately doubles pareidolic face detection rates. Interestingly, adding animal faces alongside pareidolic faces (30\%/70\% split respectively) can improve performance over fine-tuning on pareidolic faces alone.}
\label{tab:results_a2p}
\vspace{-.15in}
\end{centering}
\end{table}
 
\begin{figure}[t]
    \centering
    \includegraphics[width=0.6\columnwidth]{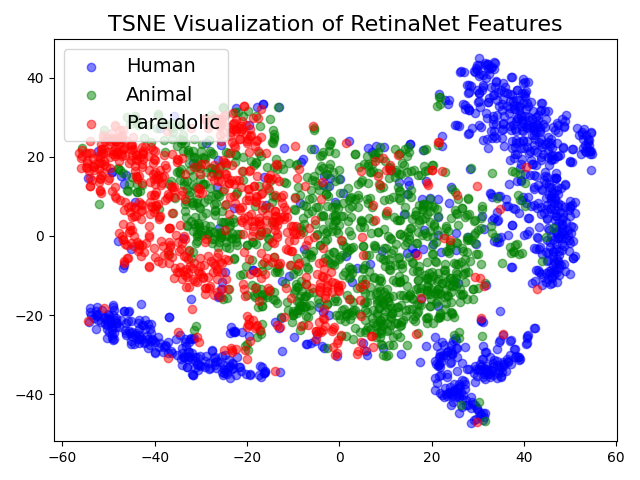}
    \vspace{-.1in}
    \caption{\textbf{Visualizing RetinaNet Representations across Datasets.} Animal+Pareidolia fine-tuned RetinaNet representations tend to group animal and pareidolic faces together. This lends evidence to the hypothesis that the perception of animal and pareidolic faces are linked. (To highlight the commonality of pareidolic animal detection we note the similarity of these points to a frog.) }

    \label{fig:feature_tsne}
\end{figure}

To understand this effect better, Fig.~\ref{fig:feature_tsne} visualizes the inner representations of this model across the three datasets (Human = WIDER FACE, Animal = AnimalWeb, Pareidolic = Faces in Things). Specifically, we extract multi-scale features from the animal and pareidolia fine-tuned RetinaNet shared feature layer before the application of the classification and regression heads. We average pool these features across the bounding box for each face and visualize them with t-SNE~\cite{hinton2002stochastic}. This figure shows that RetinaNet's representations of animal and pareidolic faces tend to cluster together and are distinct from its representations of human faces. This lends evidence to the relative similarity of pareidolic and animal faces compared to human faces. We also reiterate that we filtered the Faces in Things dataset to avoid images of real animals. 

\section{Modeling Pareidolia}
\label{sec:model}

Though many prior works have measured pareidolia, there has yet to be a simple mathematical model that describes the high-level structure of this phenomenon. In this section we provide two simple formal models of pareidolia and show that they both exhibit a testable prediction: the existence of a peak in pareidolic face detection as a function of an image's complexity. Section \ref{sec:empirical_peak} presents experimental evidence of this ``pareidolic peak'' in both humans and machines.

\subsection{Gaussian Model of Pareidolia}
\label{sec:gaussian}

 A model of pareidolia needs to describe two processes:  (1) the random process that generates candidate images, and (2) the face detection process which determines when an image is pareidolic. We begin with a simple Gaussian model for each. We model the image generation process as a sum of independent normal modes, each contributing a zero-mean Gaussian of a specified variance, multiplied by the mode image $y_i$. For example, as in \cite{duda2012pattern,Turk91}, these modes could be the principal components of a mean-subtracted image dataset. In this setting the generated image, $y$, is a weighted sum of the normal modes:
\begin{equation}
\vec{y} = \sum_i n_i \vec{y}_i \quad \textrm{ where, } \quad n_i \sim N(0, \sigma_i)
\end{equation}
To model our face detection process we capture the intuition of matching an image to a template image and note that this can be generalized to distributions of template images. In particular, the target pareidolic image is represented as a vector, $\vec{a}$, of statistically independent target coefficients, $a_i$, for each mode. The probability that this mode contributes towards the face detection, $P(a_i)$, is the probability of detecting the pareidolic value, $a_i$, at the $i$th mode. We assume a Gaussian detection process: $P(a_i|y_i) \sim N(a_i, \gamma_i)$. Because each mode's coefficient is a zero-mean Gaussian distribution, $P(y_i) \sim N(0, \sigma)$, we have:
\begin{eqnarray}
  P(a_i) & = & \int_{y_i} P(a_i,  y_i) \mbox{d}y_i \\
  & = & \int_{y_i} P(a_i | y_i ) P(y_i )  \mbox{d}y_i \\
  & = & \frac{1}{2 \pi \gamma_i \sigma_i}\int_{y_i} e^{-\frac{(y_i-
        a_i)^2}{2 \gamma^2_i}}
        e^{-\frac{y_i^2}{2 
        \sigma^2}}  \mbox{d}y_i 
        \label{eq:modepareidolia}
\end{eqnarray}
Note that $\sigma_i^2$ is the variance of the random process generating the pareidolia, while $\gamma_i^2$ is the variance of the likelihood term --- how far a mode is allowed to vary from the target mode value before it stops looking like the target image, $\vec{a}$.

We can complete the square in Eq.~\ref{eq:modepareidolia} to write the product of Gaussians in $P(a_i)$ as a single Gaussian.  Integrating that Gaussian over all possible observations $y_i$ gives the probability of finding the pareidolic value $a_i$ from mode $i$:
\begin{equation}
  P(a_i)  = \frac{1}{\sqrt{2 \pi (\gamma_i^2 +  \sigma_i^2)}}
                 e^{-\frac{a_i^2}{2 ( \sigma_i^2 + \gamma_i^2)}}
\end{equation}

\begin{figure}[t]
\centering
\vspace{-.1in}
\includegraphics[width=.8\linewidth]{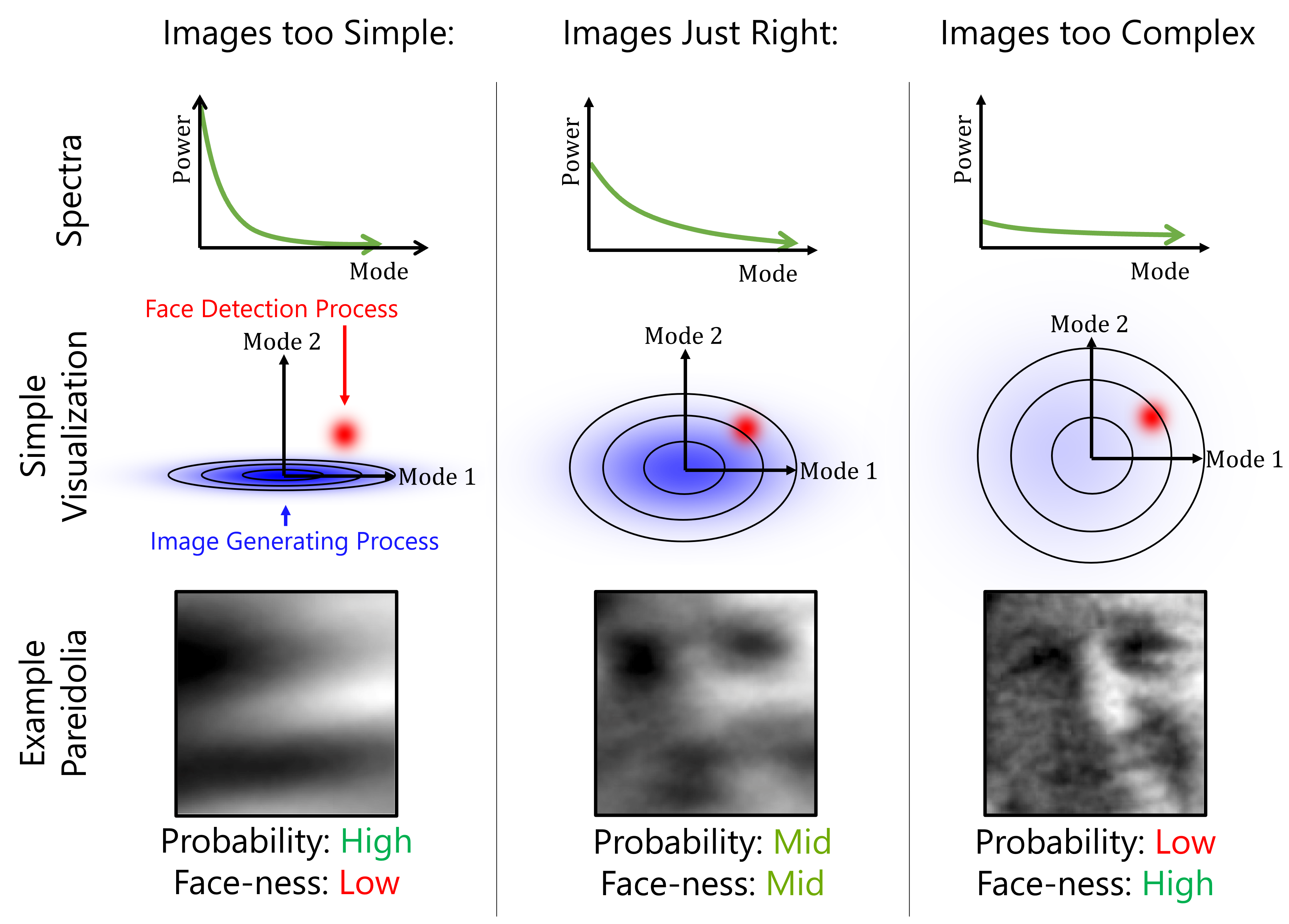}
\vspace{-.1in}

\caption{Illustration of the proposed Gaussian model for pareidolia with three example generating distributions. To make pareidolia likely, the generating distribution needs a proper distribution of spatial frequencies. A process with too few spatial frequencies (left) is likely to only generate weak face-like details (``face-ness'': low). In contrast, with too many frequencies (right), faces can be modeled with exquisite detail (``face-ness'': high), but the likelihood of drawing any particular desired combination become vanishingly small. The most likely pareidolic images form when the generating distribution has the right spectrum (middle), enabling reasonable faces to emerge with reasonable likelihood. In other words, this model predicts that pareidolic faces will match the low frequencies of faces but differ in the higher frequency details.}
\label{fig:modePareidoliaProb}
\end{figure}

\subsection{Predicting Peak Pareidolia}

For a given mode's detection variance, $\gamma_i^2$, and target mode coefficient, $a_i$, 
Eq.~\ref{eq:modepareidolia} allows us to find the optimal mode variance to generate pareidolia, i.e., to maximize $P(a_i)$. Unfortunately, we seldom have the flexibility to design a random process one mode at a time.
But we may have the option to select between image generation processes that have different numbers of modes, $M$.
Since each mode is independent, the probability of a pareidolic detection of the target object template is the product of detecting the target coefficient for each of the M modes:
\begin{equation}
  P(a)  =  \prod_i^M P(a_i)     
\label{eq:analytic3}
\end{equation}

\begin{figure}[t]
\begin{minipage}[c]{0.48\textwidth}
\centering
\includegraphics[width=\columnwidth]{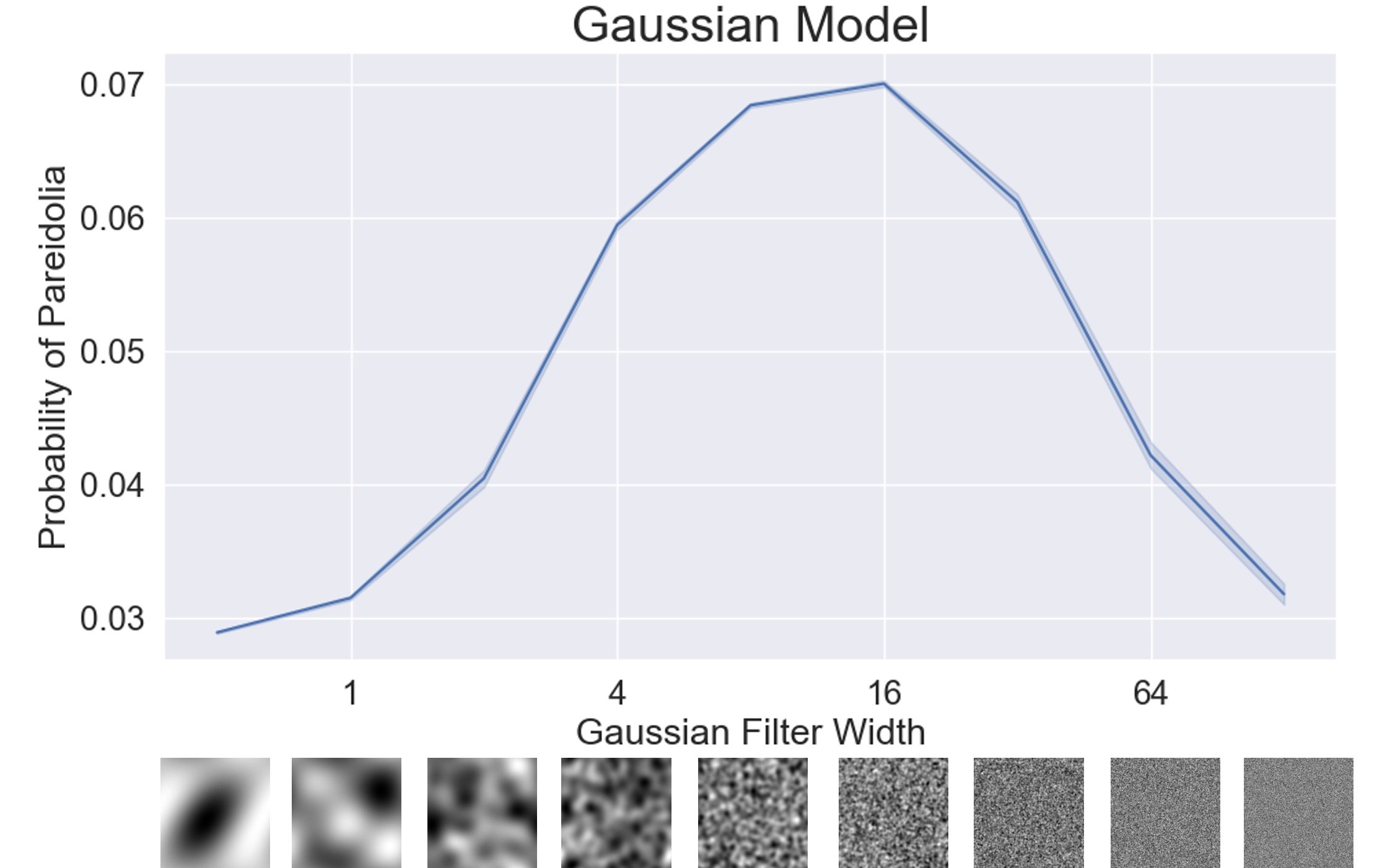}
\vspace{-.2in}
\caption{Probability of pareidolia (Eq.~\ref{eq:analytic3}) in the Gaussian model ($\sigma=10$) across images with different spatial frequency distributions (Sec. \ref{sec:empirical_peak}). This assumes spatial frequencies are uncorrelated and thus underestimates the probability of pareidolia, however peak pareidolia is still present.
}
\label{fig:gaussPeakPareidolia}
\end{minipage}%
\hfill
\begin{minipage}[c]{0.48\textwidth}
\centering
\vspace{-.15in}
\includegraphics[width=\columnwidth]{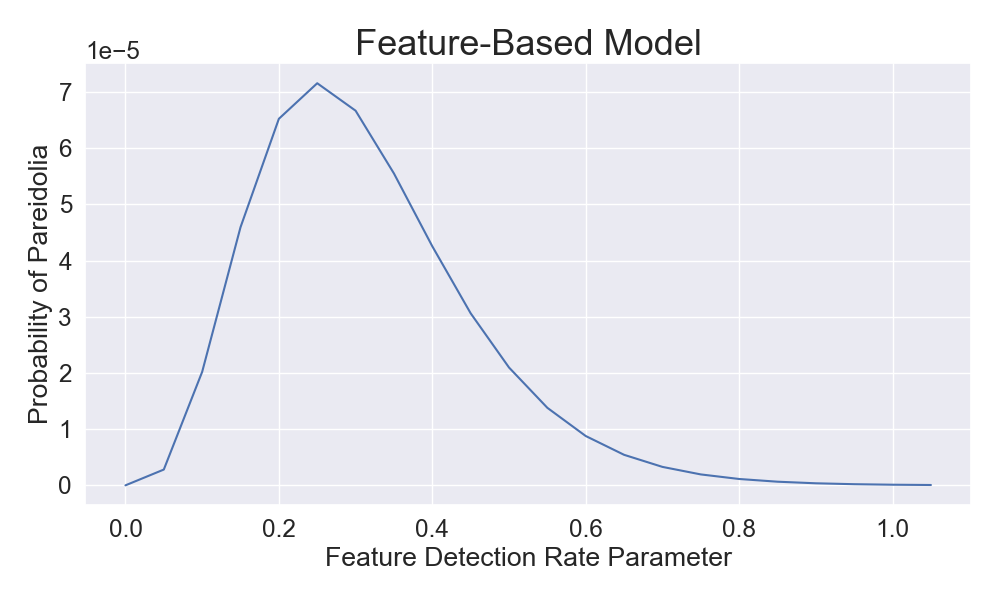}

\caption{Probability of pareidolia under the feature-based example of Eq.~(\ref{eq:featureProb}) as a function of the rate of feature detection, $\lambda_i = \lambda$ for all $i$, within the random images.  Note the low probability of pareidolia for both feature-free ($\lambda \rightarrow 0$) and feature-rich ($\lambda \gg 0$) random images.}
\label{fig:featureProb}
\end{minipage}
\end{figure}

We plot some predictions of our Gaussian model in Fig.~\ref{fig:gaussPeakPareidolia} for a target template with a $\frac{1}{f}$
power spectrum (standard deviation of each mode inversely proportional to mode number) on noise images of varying complexity. We note the existence of a peak in pareidolic detection probability for random image generation processes with a mid-range number of spatial modes, as measured by the width of Gaussian that modulates power in Fourier space. Too few modes in the random generation process, and no image will ever have enough complexity to render the target well. Too many modes and pareidolia becomes unlikely because so many modes need to match the desired target values. Each added mode multiplies the pareidolia probability by another small factor. In between, there is what we call {\em peak pareidolia}. As the detection value, $\sigma$, becomes more stringent (smaller) the peak pareidolia value occurs at a larger number of modes and becomes less probable. We illustrate this effect in Figure \ref{fig:modePareidoliaProb}

\subsection{Higher-Level Feature Model of Pareidolia}

The Gaussian model for pareidolia above lays out important aspects of pareidolia, but relies on a naive model of object detection, the squared distance from a template image. We assume that a more realistic model of human perception would incorporate higher-level features and introduce a still simple, yet more realistic, feature-based model.

We assume that the detection of an object requires particular features to be detected in certain spatial regions, e.g. an eye in the top left and right, a nose in the center, and a mouth in the bottom. Such an approach has been used in computer vision object detection algorithms, e.g.~\cite{Yuille1991,Weber2000,Felzenszwalb2010}.  Any given object template has some number of regions, $R_i$, indexed by $i$, within which a given feature, $F_i$, must be detected.  The other features, $F_{j \ne i}$, should not be detected in region $i$. For a given random image where we hope to detect pareidolia, we assume that feature existence is a spatial Poisson process. In this process, the probability of $n$ feature instances for any given feature $i$ over some area $B_i$ is

\begin{equation}
P(n_i) = \frac{(\lambda_i |B_i|)^{n_i}}{n_i!} e^{-\lambda_i |B_i|}
\end{equation}

To detect a pareidolic instance of the object
template, we must detect one feature of the correct type, $F_i$, in each region $i$ of the face template, and zero features of the wrong type, $F_{j \ne i}$ in each region $i$. Assuming independence of the feature detections, and for simplicity setting all the feature detection rates to be the same, $\lambda_i = \lambda$, and all the template areas to be the same, $B_i = 1$, we have for the probability, $P(O)$,
of pareidolic detection of object $O$:
\begin{equation}
  P(O)  =  \prod_i^M \lambda^1 e^{-\lambda (M-1)} 
\label{eq:featureProb}
\end{equation}

For the case of $M=4$, a simple detection model for two eyes, a nose, and a mouth, we have $P(o) = \lambda^4 e^{-16 \lambda}$, which is plotted in Fig.~\ref{fig:featureProb}.
In this feature-based object detection model, we also find
the existence of ``peak pareidolia''. Again, it is governed by a parameter describing the complexity of the random image, in this case a Poisson process rate parameter, $\lambda$, that governs the probability of a feature detection per unit area. For too low a rate, the model doesn't generate enough features to satisfy the object template, for too high a rate, the probability of seeing only the right features in just the right places becomes very small. In between is the most probable rate for pareidolia.

\subsection{Measuring the Pareidolic Peak in Humans and Machines}
\label{sec:empirical_peak}

\begin{figure*}[t]
    \centering
    \vspace{-.1in}
    \includegraphics[width=\columnwidth]{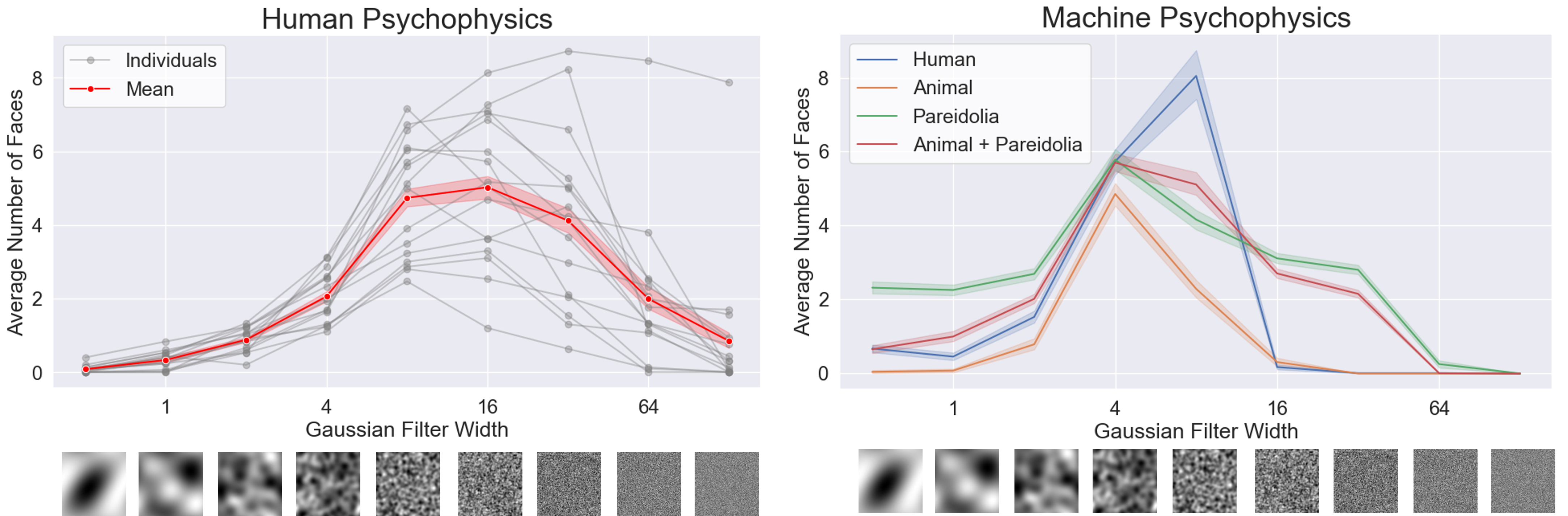} 
    \vspace{-.2in}
    \caption{\textbf{Measuring Peak Pareidolia.} Left: Subjects were asked how many faces they see in each noise image. We plot the average number of faces detected as a function of noise frequency (examples on x-axis), the mean over all subjects and its 95\% confidence interval in red. Right: average number of faces detected by our fine-tuned models. This reveals the ``peak pareidolia'' effect predicted in Section \ref{sec:model} across humans and machines.}
    \label{fig:gaussPsychophysics2}
\end{figure*}

Both mathematical models of Section \ref{sec:model} predict the existence of a peak of pareidolic face detection as a function of image complexity. We show the existence of this pareidolic peak in both humans and machines. In particular, we perform a psychophysics experiment where human subjects view noise images of varying complexity and report how many pareidolic faces they saw in each image, from zero to nine. Campbell \cite{campbell1983much} demonstrated that a 12x12 array of random, binary squares is sufficient to evoke human and animal faces. We generate noise images of varying complexity by randomly sampling Fourier coefficients and modulating these coefficients with a zero mean $\sigma^2$ variance Gaussian in Fourier space. We show some samples of these images on the x-axis of Fig.~\ref{fig:gaussPsychophysics2}. Intuitively, the Gaussian envelope in frequency space filters out most frequencies higher than $\sigma$ after applying an inverse Fourier transform. We detail our image stimuli creation method in the Supplement.

We find that humans exhibit the model-expected peak pareidolia, with a maximum number of faces detected at a frequency filter width of 16 (Fig.~\ref{fig:gaussPsychophysics2}, left). The existence of a pareidolic peak at or near this filter width is consistent among \textit{all} subjects even for those that reported fewer faces overall. Although response time did decrease slightly at higher frequencies, it did not fall off completely at the highest frequency levels, indicating that fewer reported faces were not the result of subjects ``giving up'' on the task. We provide additional details and analysis of this experiment in the Supplement.

Finally, we evaluate our fine-tuned models from Section~\ref{sec:exp} on the same images to test whether machines also exhibit peak pareidolia (Fig.~\ref{fig:gaussPsychophysics2}, right). In particular, we showed the models $5,000$ sampled noise images of varying frequency levels and counted the number of face detections they make with confidence $>10\%$. We find the same characteristic ``pareidolic peak'' where models detect the most faces in medium-complexity images.

\section{Conclusion}
\label{sec:conclusion}
We have taken initial steps towards the mathematical modelling of pareidolia and build a richly annotated dataset of images for face pareidolia. We showed through experiments on modern face detectors that detecting animal faces may partly explain the emergence of pareidolia in a complex vision system. 
The Faces in Things dataset can help the community address other questions about how and why pareidolic behavior emerges, a hallmark of humans' robust recognition system. 
We hope that our findings and dataset will spark further study of pareidolia and its potential use to improve computer vision systems. 

\subsection*{Acknowledgments}
We would like to thank Karen Hamilton for her hundreds of hours of annotations for the Faces in Things dataset.
We also thank Abhishek Dutta who created the VGG Image Annotation tool, for his kind generosity to support our project.

We would like to thank the Microsoft Research Grand Central Resources team for their gracious help performing the experiments in this work. Special thanks to Oleg Losinets and Lifeng Li for their consistent, gracious, and timely help, debugging, and expertise. Without them, none of the experiments could have been run. 

This material is based upon work supported by the National Science Foundation Graduate Research Fellowship under Grant No. 2021323067. 
Any opinion, findings, and conclusions or recommendations expressed in this material are those of the authors and do not necessarily reflect the views of their employers, or the National Science Foundation.

This work is supported by the National Science Foundation under Cooperative Agreement PHY-2019786 (The NSF AI Institute for Artificial Intelligence and Fundamental Interactions, http://iaifi.org/) and the CSAIL MEnTorEd Opportunities in Research (METEOR) Fellowship.

Research was sponsored by the United States Air Force Research Laboratory and the United States Air Force Artificial Intelligence Accelerator and was accomplished under Cooperative Agreement Number FA8750-192-1000. The views and conclusions contained in this document are those of the authors and should not be interpreted as representing the official policies, either expressed or implied, of the United States Air Force or the U.S. Government. The U.S. Government is authorized to reproduce and distribute reprints for Government purposes notwithstanding any copyright notation herein.

The authors acknowledge the
MIT SuperCloud~\cite{reuther2018interactive} and Lincoln Laboratory Supercomputing Center for providing HPC resources that have contributed to the research results reported within this paper.

\bibliographystyle{splncs04}
\bibliography{main}



\appendix

\section{Appendix}

\subsection{Additional Information on Frequency-Dependent Noise Generation}
\label{sec:noise-details}

To generate noise of different frequencies for our experiments we leveraged the fact that low frequency information is localized close to the origin in the Fourier transform of an image. To this end, we can generate a random noise images by randomly sampling images in the Fourier space, filtering them, and transforming back to image space. Specifically, we modulate a random Fourier spectra by a Gaussian centered at 0 with a variable width. The width controls the frequency of the noise created. Larger width images let more frequencies pass through in Fourier space, and the resulting image has higher frequency patterns.

\begin{figure}[ht]
\centering
\includegraphics[width=0.5\columnwidth]{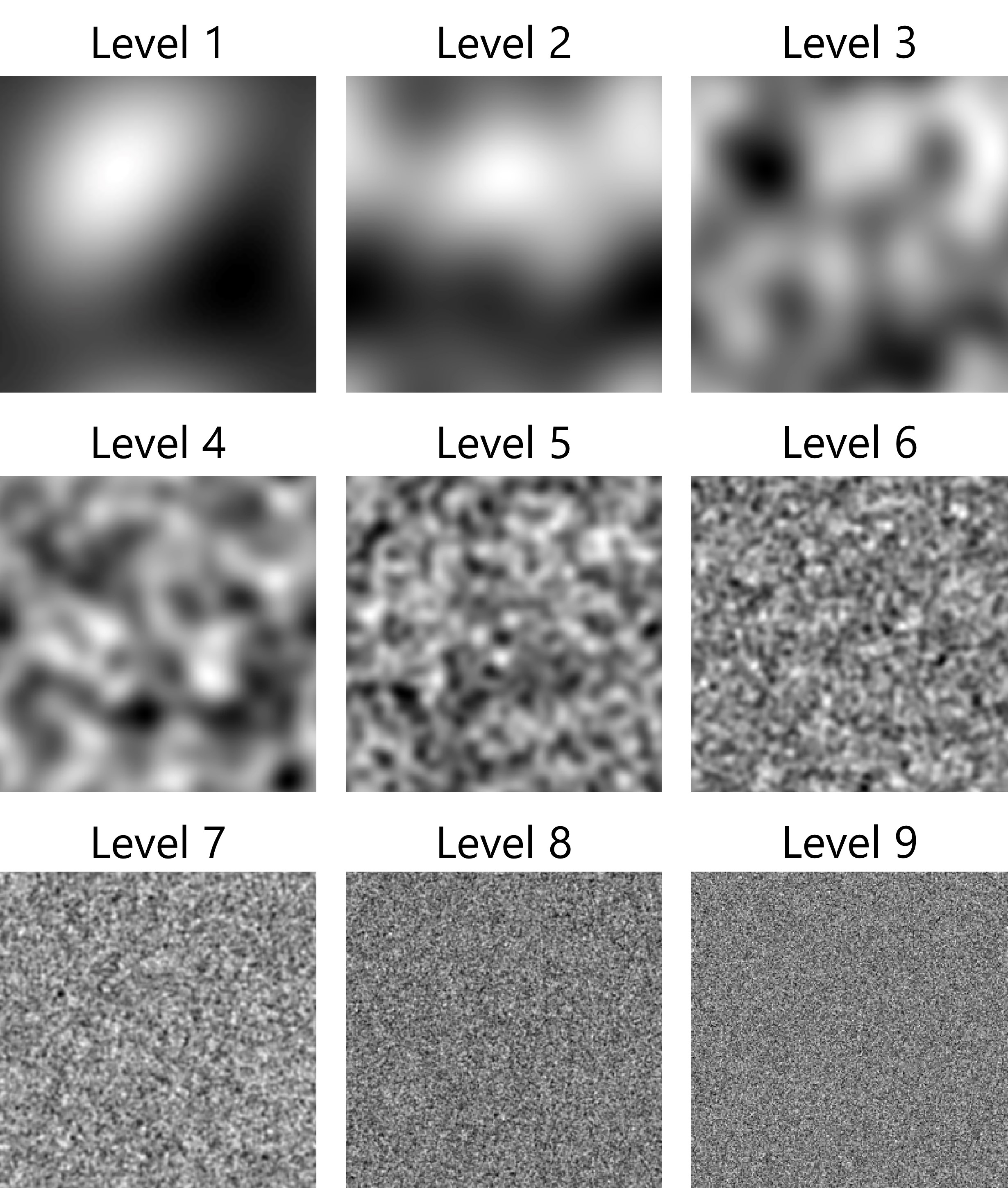}
\caption{Different frequency noise levels used in human experiments from Fig.~\ref{fig:gaussPsychophysics2}.}
\label{fig:noise_images}
\end{figure}

For the noise levels shown in Fig.~\ref{fig:noise_images} we use a width of $2^{level-2}$. We include the following pseudo-code to be precise:

\begin{lstlisting}[language=Python]
import numpy as np
from scipy.fftpack import fft2, fftshift, ifft2
def generate_noise(width, size):
    modes = np.randn(size, size)
    dft = fft2(modes)
    gauss = fftshift(np.exp((-xx ** 2 - yy ** 2) / (2 * width ** 2)))
    return np.real(ifft2(dft * gauss))
\end{lstlisting}



\subsection{Human Psychophysics Experiment Setup}
\label{sec:appendix:humanpsychophysicssetup}

In the main experiment, 14 subjects (6 female, 8 male) were shown noise images filtered as described in \ref{sec:noise-details} at resolution 1024x1024. At each of 9 filter widths, 10 random noise pulls were created, making 90 unique images. Each unique image was repeated 3 times, for a total of and 270 presentations per subject. The images were shown in random order, which was different for each subject. Subjects were split into two groups of 7 subjects, with the two groups view different sets of 90 unique images. Experiments were performed in PsychoPy, and the experimental code both for generating and displaying experimental stimlui is available at \href{https://github.com/vdutell/pareidolicNoise}{https://github.com/vdutell/pareidolicNoise}.

Subjects were seated in a dimly-lit room in front of a laptop with screen resolution 2500x1664, with the 1024x1024 image subtending the entire screen vertically, with grey padding on the horizontal edges (except for 2 additional subjects in the control experiment described below, where image subtended half screen height). Subjects were instructed to sit at a comfortable distance from the screen (approximately 30 inches), and were asked to count the number of faces seen in each noise image, and report the number from [0-9] on the laptop keyboard, reporting 9 if they saw 9 or more. There were no time-outs, no response feedback, and subjects were instructed to self-pace. The experiment took approximately one hour to complete. Subjects were told that there were no ``correct'' answers, and instructed to count any face, animal or human as long as they ``felt they saw some kind of face''. Subjects were allowed to take breaks as needed. 

All participants provided informed consent prior to participation, in compliance with the Common Rule (45 CFR 46), and this study was assessed as exempt from review by the Institutional Review Board, pursuant to 45 CFR 46.101(b)(2). Participants took between 45-90 minutes complete the study and were paid $\$20$ for their participation. Subjects were not excluded for having corrective lenses, but confirmed to be able to see the screen clearly. One subject reported having vision issues beyond using corrective lenses (possible prosopagnosia, see below).

For the analysis, trials were removed where subjects responded in less than 100 milliseconds (likely to be a mistake), as well as trials where the subject took more than 2 minutes to respond (likely to be taking a break). No subjects were removed due to outlying or erroneous data. Subjects' time to completion varied from approximately 45-90 minutes. 

We analyze the effect of different psychophysical conditions with mixed effects ANOVA for all 16 subjects using image seed group, gender, image field of view (FOV) as between-subject factors, and Gaussian filter width as a within-subject factor. Uncorrelated p-values are reported for the two ANOVA analyses (image seed and gender). FOV ANOVA is omitted due to unbalanced sample sizes (14 and 2). No significant differences were found in responses for the two subject groups that were shown image sets from two different random seeds ($p>0.2$, Fig. \ref{fig:pareidoliagroupsrt}, Left). This indicates that peak pareidolia is not an artifact of a `lucky draw' from the random image set generated. 

In addition to the peak in number of faces reported, we also found that subject's response time (RT) mirrored a similar curve for low to medium filter widths. However, for the largest filter widths shown, the RT curve did not fall off as steeply as the pareidolia curve (Fig. \ref{fig:pareidoliagroupsrt}, Right). This indicates that subjects took time to look for faces in high frequency data and did not simply `give up' on the task.

\subsection{Additional Human Psychophysics Results}
\label{sec:appendix:humanpsychophysicsresults}

\begin{figure}[H]
\centering
\includegraphics[width=1\columnwidth]{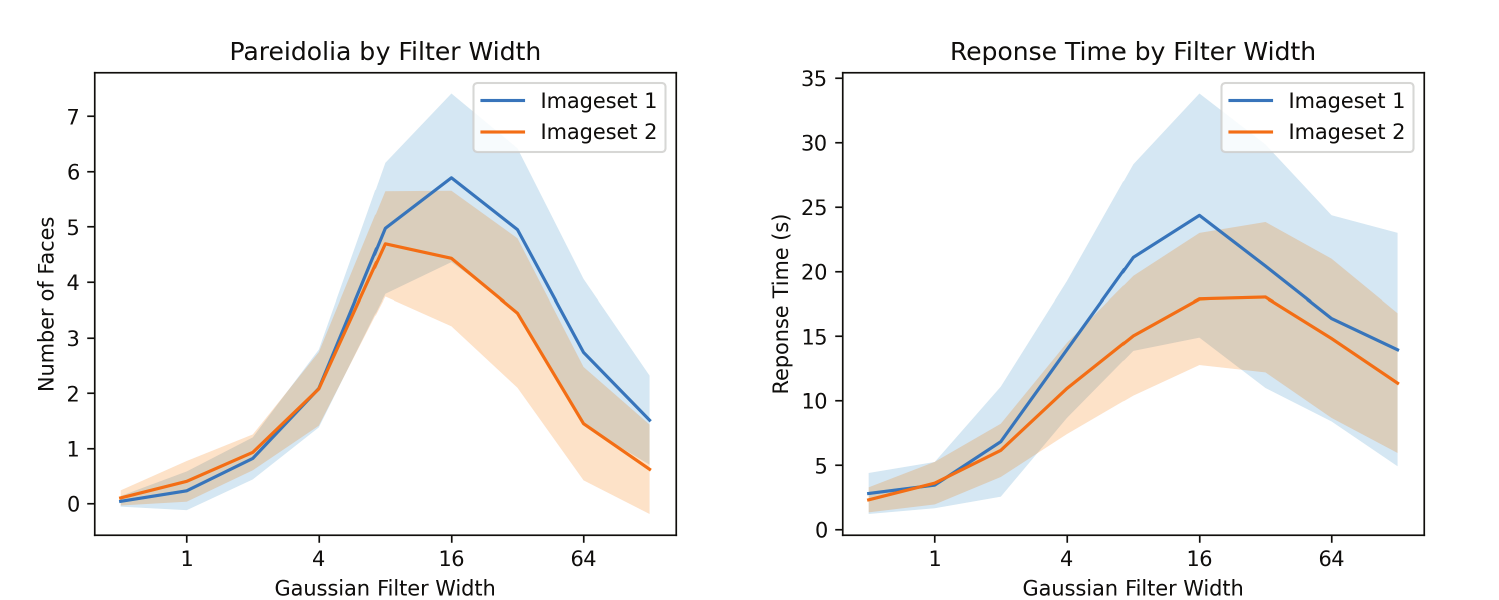}
\caption{Left: Pareidolia reported for two subject groups with different random seed images shows no significant difference between groups. Right: Response time (RT) mirrors number of faces for small to medium filter widths, but does not fall off as sharply.  Plots report $\pm 1$ standard deviation.}
\label{fig:pareidoliagroupsrt}
\end{figure}

 \begin{figure}[H]
\centering
\includegraphics[width=1\columnwidth]{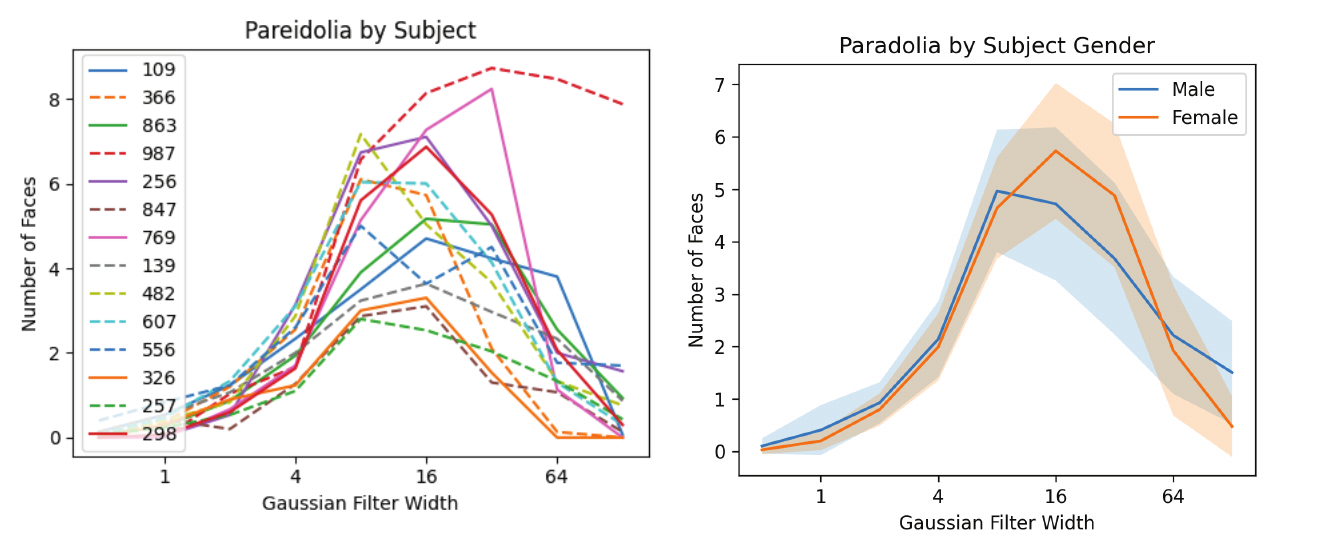}
\caption{Left: Pareidolia by subject. Right: Female and Male subjects both demonstrate peak pareidolia, and do not show significant differences. Plot reports $\pm 1$ standard deviation.}
\label{fig:gender_differences}
\end{figure}

 Furthermore, no significant differences were found between male and female subjects ($p>0.8$, Fig \ref{fig:gender_differences}, Right). We note that previous work found females have stronger pareidolic-like neural responses \cite{proverbio2016women}. Interestingly, one subject self-reported having difficulty recognizing face identity (possible undiagnosed prosopagnosia), yet still demonstrated peak pareidolia, though reported fewer faces than most other subjects (Fig. \ref{fig:pareidoliagroupsrt}, Left, Subject ID 326).

 \begin{figure}[H]
\centering
\includegraphics[width=\columnwidth]{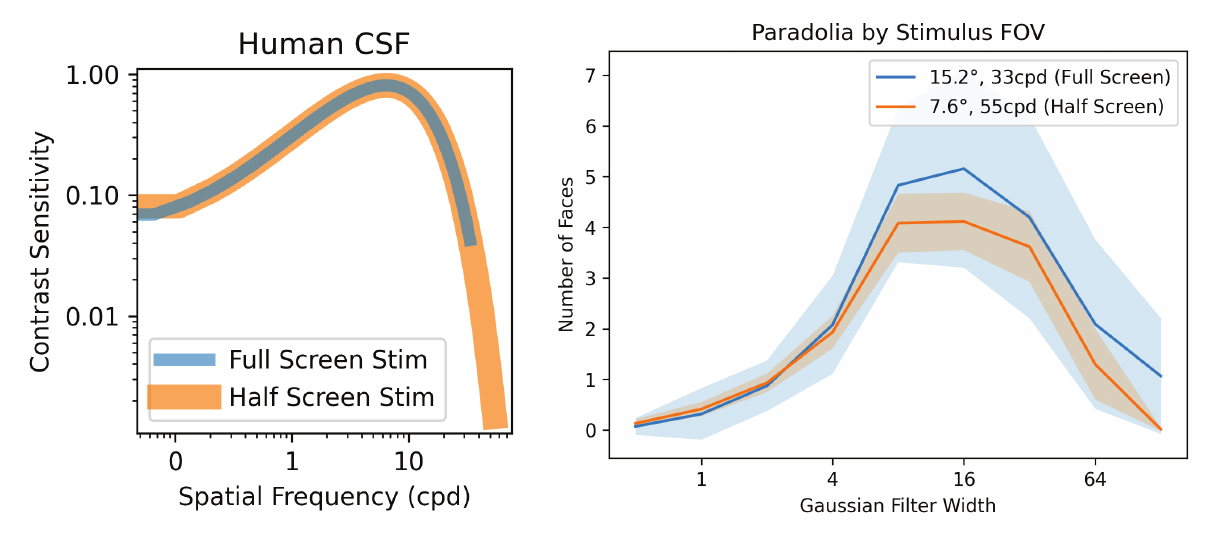}
\caption{Left: Human Contrast Sensitivity Function for 1024x1024 image shown at full screen height Field of View (full FOV, blue), and at half size (half FOV, orange). Right: Peak pareidola measured in humans for stimuli viewed at full and half FOV. Full Screen data for 14 subjects, Half screen data for two subjects. Plot reports $\pm 1$ standard deviation.}
\label{fig:CSF_FOV}
\end{figure}

Despite finding similar peak pareidolia between humans and and machines, one unaccounted difference between their visual systems is in the Contrast Sensitivity Function (CSF). The CSF describes the attenuation of spatial frequency sensitivity for humans at frequencies below and above around 10 cycles per degree (cpd). We calculate the CSF for the range of our experiment assuming the viewing distance of 30 inches, and the scotopic-mesiopic viewing conditions of our setup using the CSF equation from  \cite{mannos1974effects}. We plot this for the full screen viewing experiment in Figure \ref{fig:CSF_FOV}, Left, blue line. 

Because our model posits that high frequencies disrupt pareidolia by making mode alignment unlikely, we explored whether the human CSF, which reduces sensitivity to high frequencies, might explain the results. To determine if the CSF had a measurable effect on the human pareidolia results in our experimental setup, we ran two additional subjects in the same experiment, but with images presented at half the screen width, and therefore with half Field of View (FOV). This increased the Nyquist frequency from 33cpd to 55cpd (screen monitor limit). The Human CSF for this range of frequencies is shown in Figure \ref{fig:CSF_FOV}, Left, orange line. We find that for the two subjects tested, results trend to similarly to the full FOV condition (Figure \ref{fig:CSF_FOV}, Right) demonstrating that peak pareidolia is not solely caused by the human CSF.  


\subsection{Additional Annotation Details}
\label{sec:anno-details}

Each image in the Faces in Things dataset is annotated by the following series of questions:

\begin{enumerate}
\item Is there a face?\\(Yes / No / Several)
\item If Yes / Several: draw bounding box over face(s)
\item Is the face difficult to spot?\\(Easy / Medium / Hard)
\item Was the face generated by accident, or by design?\\(Accident / Design)
\item What emotion does the face show?\\(Neutral / Happy / Sad / Surprised / Angry / Disgusted / Scared / Other)
\item What does the face most resemble?\\(Human-Baby, Human-Child, Human-Adult, Human-Older, Alien, Animal, Cartoon, Robot, Other)
\item What gender do you think the face is?\\(Neutral / Female / Male)
\item Is this example of pareidolia amusing?\\(No / Somewhat / Yes)
\item How common is this type of face pareidolia?\\(Uncommon / Somewhat / Common)
\end{enumerate}

For consistency, a single annotator was tasked with annotating raw data until a set of five thousand face-containing images had been collected. 
Data was then manually checked by the authors to correct errors, confirm the reasonableness of the annotations and to flag any faces that were considered unsafe for viewing. Duplicates were automatically detected and removed by thresholding the similarity between or DINOv2 class tokens at $0.85$.
We note that the subjectivity of this task and that the annotations represent a biased view of the dataset from a single annotator's perspective.
While it would be interesting to annotate the same data with multiple annotators to build a distribution of answers and model this subjectivity, it was outside the scope of our current project and we leave it as a direction for future work.

\newpage

\subsection{Additional Average Face Renderings}

\begin{figure}[h]
    \centering
    \includegraphics[width=0.5\columnwidth]{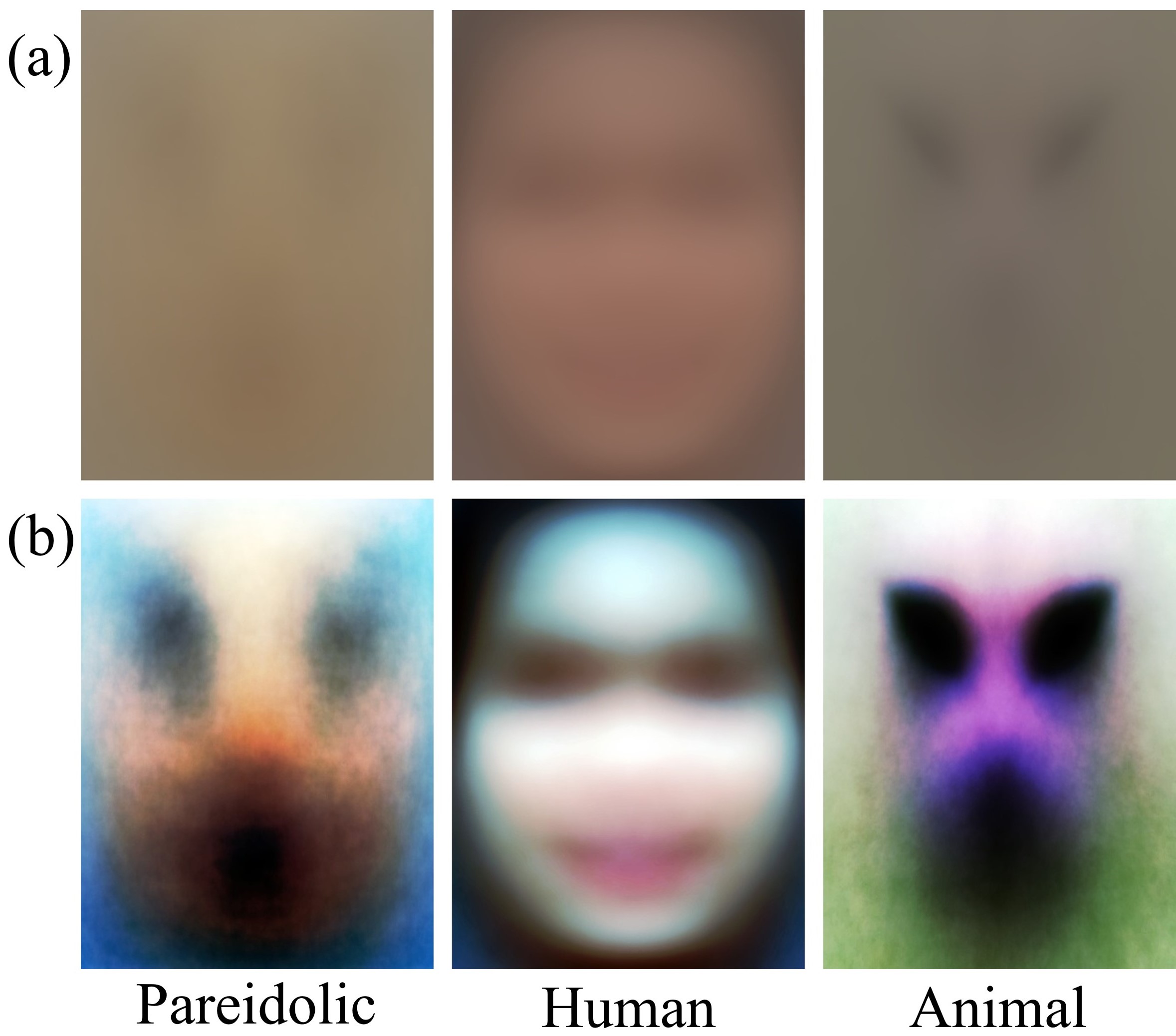}
    \caption{\textbf{The Appearance of an Average Pareidolic Face.} Shown here are the (a) raw average and (b) per-channel histogram-equalized average images for registered pareidolic faces (our Faces in Things dataset), human faces (samples from the WIDER FACE dataset~\cite{yang_wider_2016}), and animal faces (AnimalWeb~\cite{khan2020animalweb}). The average pareidolic face, while less distinct than human or animal, has surprisingly clear eye, nose, and mouth features, and vertical symmetry.}
    \label{fig:average_faces_raw}
\end{figure}

\newpage

\subsection{Average Faces across Different Conditions}
\label{sec:mean-viz}

\begin{figure}[h]
\centering
\includegraphics[width=\columnwidth]{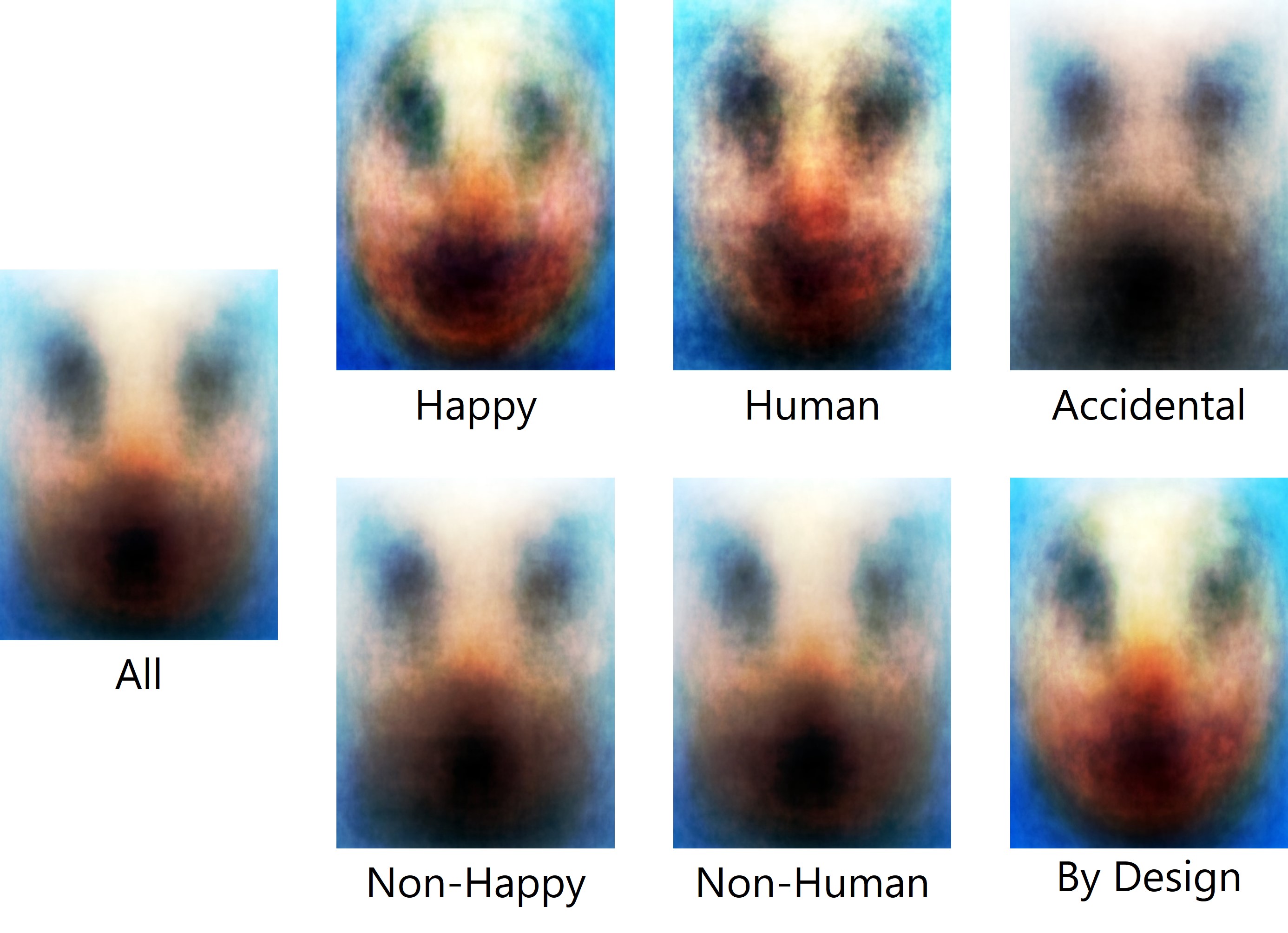}
\caption{Average faces from the Faces in Things dataset that fit certain label criteria.}
\label{fig:happy_or_not}
\end{figure}

Out of curiosity, we plot in Fig.~\ref{fig:happy_or_not} the histogram-equalized averages for faces in our dataset that have been classified as Happy (31\% of the data) or otherwise (Neutral / Sad / Surprised / Angry / Disgusted / Scared / Other).

\subsection{Why focus on Pareidolia?}
Many computer vision researchers are inspired by the human visual system and its ability to robustly recognize patterns in the world. Face pareidolia is fascinating because it is a human visual representation phenomenon that is not well understood. Our dataset, models, and experimental analyses shed light on how and why it might arise. These contributions may help the community to: better understand and harness human visual attention (which is drawn towards face-like objects), reduce pareidolic false positives in face detectors, help designers avoid or create pareidolia, improve pareidolic animation, and create systems that understand how we perceive the world. 

\newpage

\subsection{Analyzing the Viola-Jones face detector.}

\begin{figure}[ht]
    \centering
    \includegraphics[width=\linewidth]{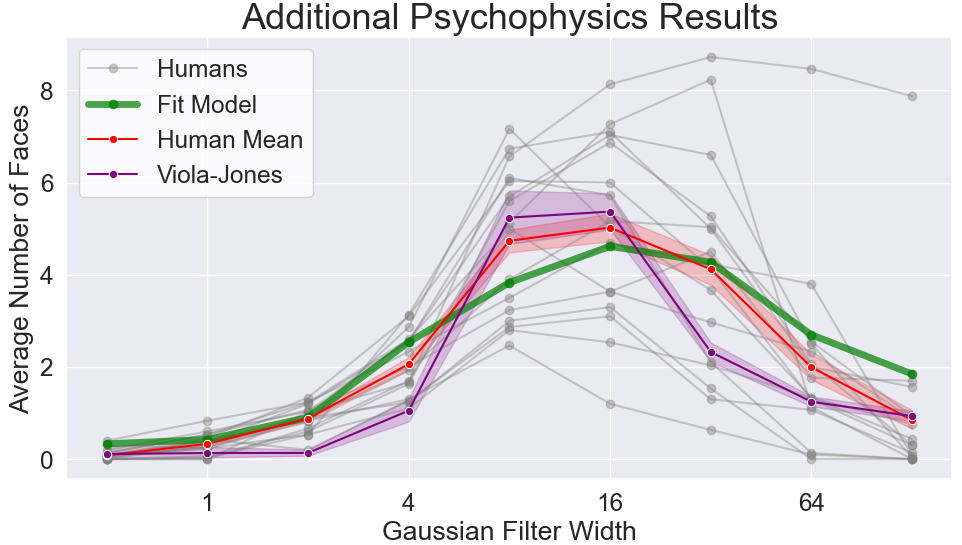}
    \vspace{-.3in}
    \caption{Fitting our Gaussian model (green) to human data. The Viola-Jones face detector also shows peak pareidolia (purple). }
    \label{fig:fit_model}
    \vspace{-.2in}
\end{figure}

We perform a set of additional experiments with the Viola-Jones face detector and find that it also displays a ``pareidolic peak'' as seen in in Figure \ref{fig:fit_model}.

\subsection{Fitting the Gaussian model of Pareidolia to human experiments.}
We fit our Gaussian model to human data in Figure \ref{fig:fit_model}. The fit parameters ($\sigma=6$) are similar to the flot of Figure \ref{fig:gaussPeakPareidolia} plots ($\sigma=10$).

\newpage

\subsection{Simultaneous Classification and Detection.}

\begin{figure}[ht]
    \centering
    \includegraphics[width=.5\linewidth]{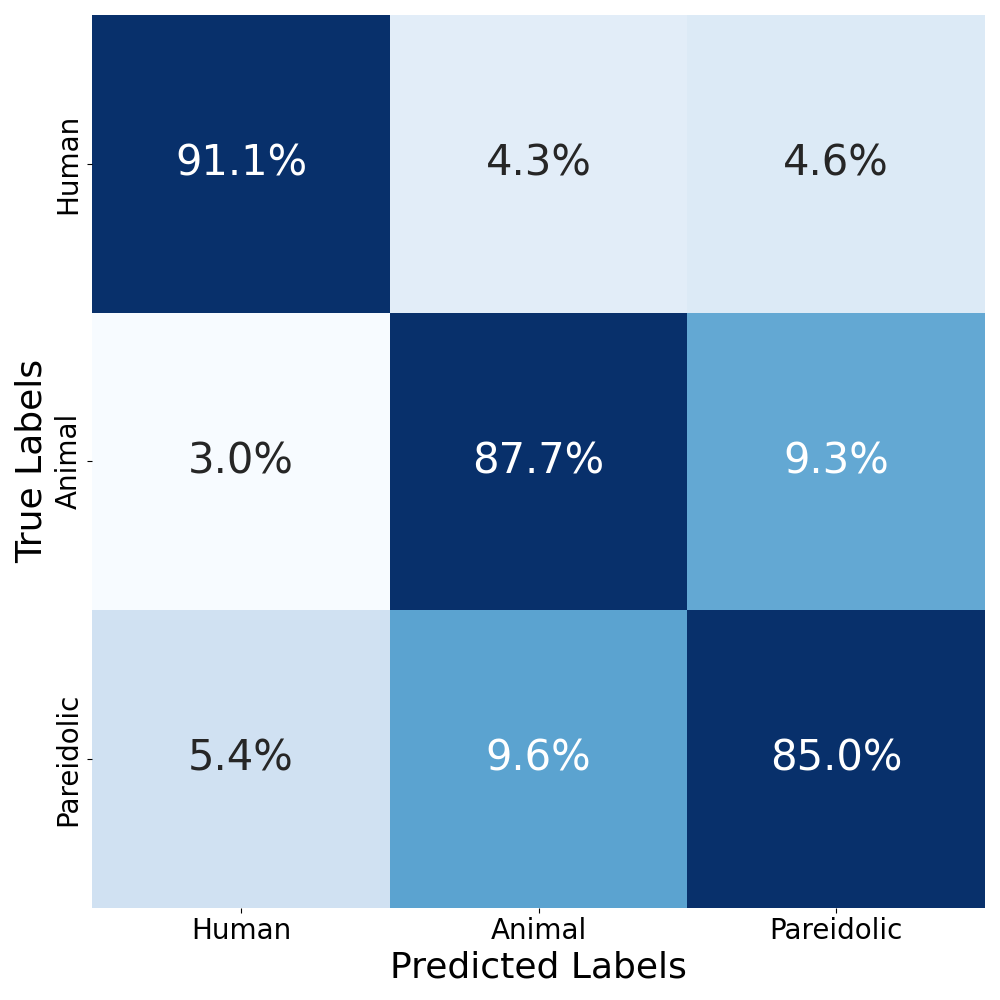}
    \vspace{-.18in}
    \caption{Simultaneous face detection and classification. Block structure shows similarity between animal and pareidolic faces.  }
    \label{fig:confusion_mat}
    \vspace{-.15in}

\end{figure}

We show a simultaneous classification analysis in Figure \ref{fig:confusion_mat}. We thank the reviewer as this further quantifies our findings that models are more likely to confuse animal and pareidolic faces with each other than with human faces. 

\end{document}


\title{Seeing Faces in Things: A Model and Dataset for Pareidolia}  

\maketitle
\thispagestyle{empty}
\appendix

\newcommand{\Rone}{\textcolor{olive}{\textbf{R1}}}
\newcommand{\Rtwo}{\textcolor{teal}{\textbf{R2}}}
\newcommand{\Rfour}{\textcolor{cyan}{\textbf{R4}}}
\newcommand{\Rfive}{\textcolor{orange}{\textbf{R5}}}


For the most up-to-date supplementary information see \href{https://aka.ms/faces-paper}{https://aka.ms/faces-paper}.